\documentclass{INTERSPEECH2023}
\usepackage{comment}
\usepackage{tipa}

\interspeechcameraready

\usepackage{booktabs,multirow,cite}

\usepackage{xcolor}

\title{How Generative Spoken Language Modeling Encodes Noisy Speech:\\Investigation from Phonetics to Syntactics}

\name{Joonyong Park$^1$, Shinnosuke Takamichi$^1$, Tomohiko Nakamura$^{1}$,\\Kentaro Seki$^1$, Detai Xin$^1$, Hiroshi Saruwatari$^1$
    \thanks{T. Nakamura is currently with National Institute of Advanced Industrial Science and Technology (AIST), Japan.}
}

\address{
  $^1$The University of Tokyo, Japan
}
\email{joonyong-park@g.ecc.u-tokyo.ac.jp, shinnosuke\_takamichi@ipc.i.u-tokyo.ac.jp}

\begin{document}

\maketitle

\begin{abstract}
We examine the speech modeling potential of generative spoken language modeling (GSLM), which involves using learned symbols derived from data rather than phonemes for speech analysis and synthesis. Since GSLM facilitates textless spoken language processing, exploring its effectiveness is critical for paving the way for novel paradigms in spoken-language processing. This paper presents the findings of GSLM's encoding and decoding effectiveness at the spoken-language and speech levels. Through speech resynthesis experiments, we revealed that resynthesis errors occur at the levels ranging from phonology to syntactics and GSLM frequently resynthesizes natural but content-altered speech.
\end{abstract}
\vspace{-1mm}
\section{Introduction}

The field of natural language processing has made significant advancements~\cite{NIPS2017_3f5ee243}, particularly in the area of language models. Studies have shown that language models can naturally learn multiple tasks even without supervision, which can enable the model to comprehend, answer questions, summarize, and translate text~\cite{devlin-etal-2019-bert,lewis-etal-2020-bart,NEURIPS2020_1457c0d6}.
Despite the success in text-level language modeling, it does not handle spoken-language expressions, which is another essential component of speech.
To resolve this problem, investigations on speech-language models through the use of extensive unlabeled speech data have been conducted~\cite{abs-1807-03748,chung19_interspeech,9054458,NEURIPS2020_92d1e1eb,9585401,Chen_2022}.



The most promising approach to this problem is generative spoken language modeling (GSLM)~\cite{lakhotia-etal-2021-generative}, which involves learning the acoustic and spoken linguistic features only from speech audio.
As shown in Fig.~\ref{fig:fig/ronbun_gslm_arch.pdf}(a), GSLM involves analyzing and synthesizing speech audio through a sequence of discrete symbols.
These symbols can be viewed as pseudo-phones, though they are not exactingly equivalent to phonemes or phones as those defined in phonetics~\cite{lakhotia-etal-2021-generative, wells22_interspeech, deseyssel22_interspeech}.
Since these symbols can be determined through training, GSLM does not require transcriptions associated with speech data, i.e., it enables text-free spoken language processing~\cite{borsos22audiolm,polyak21speechresynthesisdiscrte,ao21speecht5}.
It also potentially expands the available speech data beyond carefully designed and annotated laboratory data to real-world data.

However, real-world data usually contain noise and sometimes include speech of non-target speakers.
This may affect the encoding performance in GSLM~\cite{gat2022robustness}, posing a challenge for GSLM in being robust to noise contamination. 
This motivated us to investigate what information GSLM encodes from noisy speech and identify which levels of spoken language, ranging from phonetic to syntactic levels, are affected by noise. 

%

We investigated the encoding and decoding effectiveness of GSLM in a noisy environment (see Fig.~\ref{fig:fig/ronbun_gslm_arch.pdf}(b)).
To do this, we conducted experiments on speech resynthesis from noisy speech using the GSLM model trained with speech without noise.
The experimental results were analyzed at the spoken-language and speech levels.
%

\vspace{-2mm}
\section{Generative spoken language modeling}

\begin{figure}[tb]
    \centering
    \includegraphics[width=0.98\linewidth]{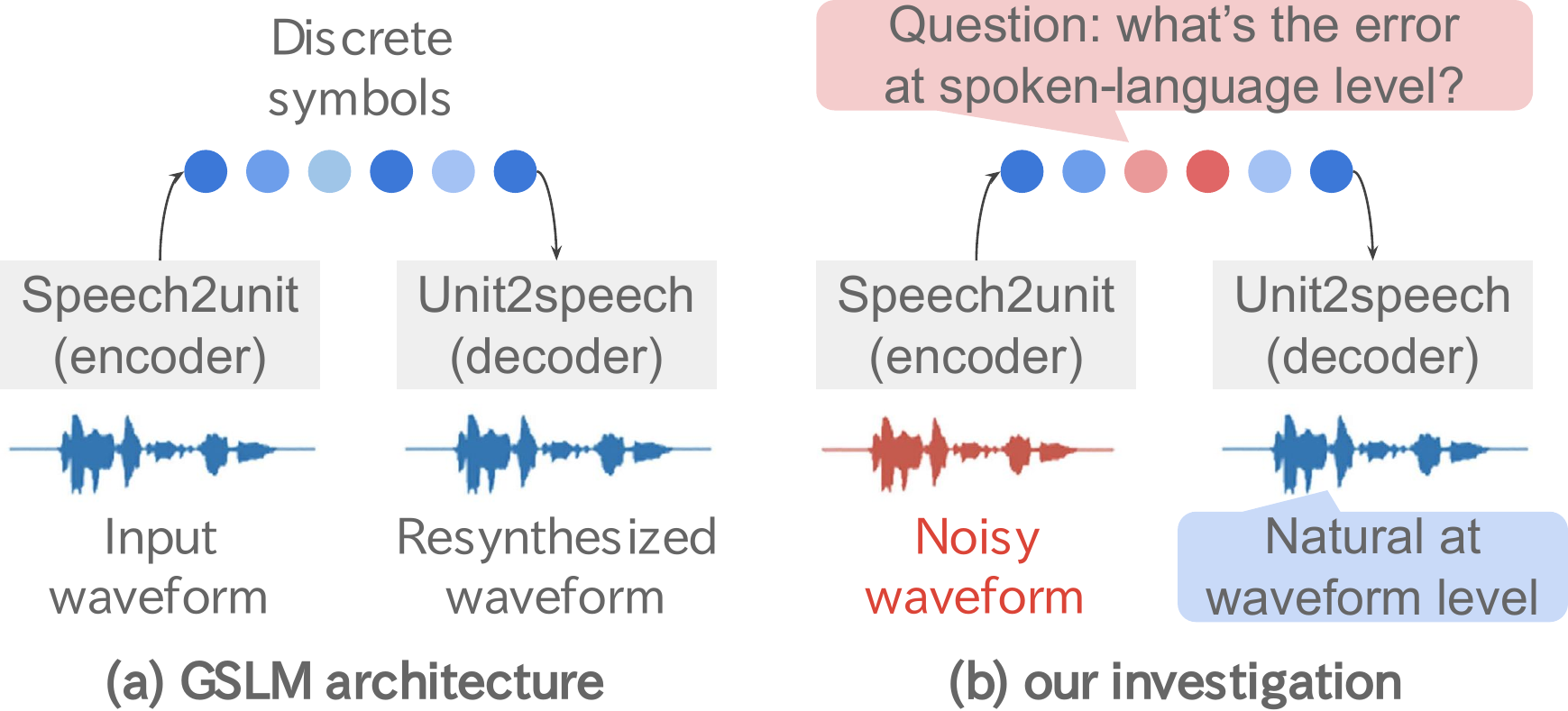}
    \caption{GSLM-model architecture and our investigation.}
    \label{fig:fig/ronbun_gslm_arch.pdf}
    \vspace{-3mm}
\end{figure}

This section outlines the architecture of the GSLM model~\cite{lakhotia-etal-2021-generative}.
Figure~\ref{fig:fig/ronbun_gslm_arch.pdf}(a) shows a schematic illustration of the GSLM-model architecture.
The GSLM model analyzes and synthesizes an audio waveform of speech through discrete symbols (units) instead of phoneme symbols.
The primary components of the GSLM model are speech2unit and unit2speech modules.
The speech2unit module first converts the audio waveform into framewise features using the encoder of a self-supervised speech representation model (e.g., contrastive predictive coding~\cite{abs-1807-03748}, wav2vec2.0~\cite{NEURIPS2020_92d1e1eb}, and HuBERT~\cite{9585401}).
It then quantizes the features with a predetermined codebook to obtain the discrete symbol sequence.
The codebook is obtained by applying $k$-means clustering to the framewise features of the training data.
The unit2speech module decodes the discrete symbol sequence into an audio waveform.
For this module, we can use the network architectures used in conventional text-to-speech synthesis models and neural vocoder models (e.g., Tacotron2~\cite{Shen2018ICASSP}).

Another important module of the GSLM model is a unit-based language model (uLM), which is omitted in Figure~\ref{fig:fig/ronbun_gslm_arch.pdf}.
It serves as a generative language model in the unit domain and enables speech generation and continuation without transcription.
However, the focus of this syudy was on speech resynthesis, which does not require the uLM.
Thus, we analyze the speech2unit and unit2speech modules.

\vspace{-2mm}
\section{Experimental analysis on noisy speech}
\subsection{Experimental setup}
We analyzed the robustness of GSLM to noise contamination through speech resynthesis experiments.
We chose HuBERT~\cite{Hsu2021ICASSP} with the codebook of 200 classes as the speech2unit module and Tacotron2~\cite{Shen2018ICASSP} as the unit2speech module, which achieved the highest performance in word error rate (WER) and mean opinion score of synthesized speech in \cite{lakhotia-etal-2021-generative}.
The GSLM model, in which the encoder is trained on the LibriSpeech corpus~\cite{7178964} and decoder is trained on the LJSpeech corpus~\cite{ljspeech17} is available via the \texttt{fairseq} toolkit~\cite{ott-etal-2019-fairseq}.


The noisy speech signals were created by adding noise signals to clean speech signals with varying signal-to-noise ratios (SNRs) from $0$ to $15$~dB with an interval of $5$~dB.
We used the \texttt{dev-clean} set of the LibriSpeech corpus (2704 utterances)~\cite{7178964} as the clean-speech signals.
The noise signals were drawn from the diverse environments multichannel acoustic noise database (DEMAND)~\cite{thiemann_joachim_2013_1227121}, which contains noise signals recorded with a 16-channel microphone array in 17 noisy environments.
We randomly chose one of the channels for each noise signal.
The number of the test dataset was $183,872$ in total.

To examine the effect of background speech, we classified the noise types into three categories along with the amount of speech contained in the noise.
Table~\ref{table:noisy_classify} shows the categorization of the noise types.
See \cite{thiemann_joachim_2013_1227121} for details of these noise types.
\textbf{L-BAB} is the collection of the noise types that contain little or no speech, i.e., low babble noise.
\textbf{M-BAB} contains the noise types including speech in less than half the signal length.
\textbf{H-BAB} is the collection of the rest, i.e., high babble noise.
Some of the background speech came from radio, television, and telephone.

\begin{figure}[tb]
    \centering
    \includegraphics[width=0.75\linewidth]{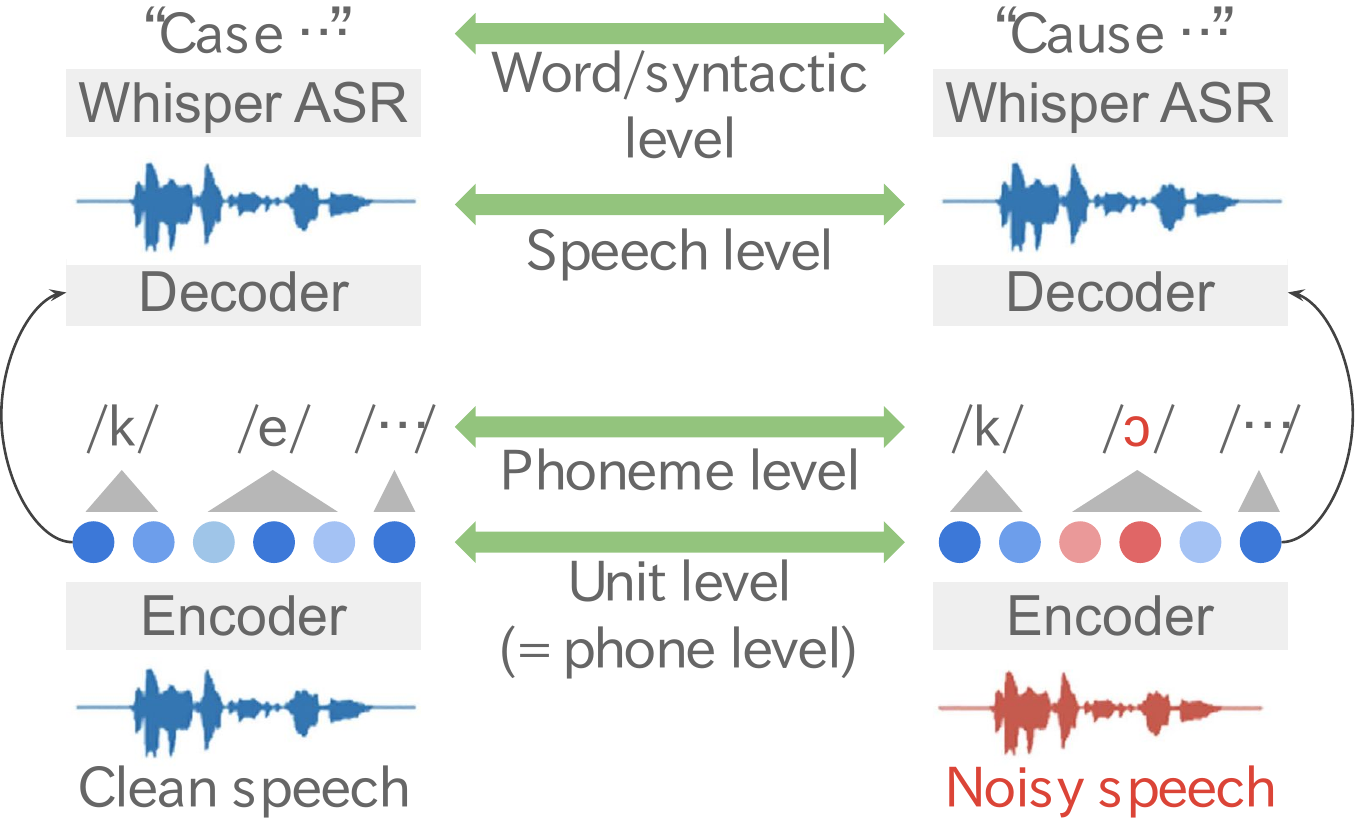}
    \caption{Levels investigated in experiments.}
    \label{fig:fig/experiment.pdf}
\vspace{-1mm}
\end{figure}

We investigated at spoken-language (Section~\ref{ssec:intelligibility}) and at speech (Section~\ref{ssec:quality}). 
The former was aimed at exploring which spoken-language content changes due to noise contamination, while the latter was aimed at finding the relation between changes in spoken-language levels and those in speech levels. Figure~\ref{fig:fig/experiment.pdf} summarizes the levels investigated.

\begin{table}[tb]
    \centering
    \caption{Categorization of noise types of DEMAND}
    \vspace{-2mm}
    \label{tb:1}
    {
        \small
        \begin{tabular}{c|c}
        \toprule
            Category & Noise type of DEMAND \\
            \midrule
            \multirow{2}{*}{L-BAB} & DKITCHIN, DWASHING, NFIELD,\\
            & NRIVER, OHALLWAY, OOFFICE, TCAR \\
            \multirow{2}{*}{M-BAB} & DLIVING, NPARK, PSTATION, \\
            & SPSQUARE, STRAFFIC\\
            \multirow{2}{*}{H-BAB} & OMEETING, PCAFETER, PRESTO, \\
            & TBUS, TMETRO
            \\ 
            \bottomrule
        \end{tabular}
    }
\label{table:noisy_classify}    
\end{table}

\subsection{Analysis at spoken language levels} \label{ssec:intelligibility}
We examined the impact of noise on the spoken language level from phonetic to syntactic levels. 

\smallskip
\noindent \textbf{Phone level:} Since the discrete symbols of GSLM can be viewed as pseudo-phones, we defined a counterpart of the WER for the discrete symbol sequences which we call the unit error rate (UER).
The UERs were computed in the same manner as the WER but using the discrete symbol sequences obtained from the clean and noisy speech as reference and hypothesis, respectively.

\begin{table}[t]
    \centering
    \caption{UERs [\%] of noisy speech}    
    \vspace{-2mm}
    \begin{tabular}{c|cccc}
    \toprule
         & \multicolumn{4}{c}{SNR}\\ 
         & 15~dB &  10~dB & 5~dB & 0~dB \\ 
        \midrule
        {L-BAB} & 13.4 & 17.1  & 22.0 & 28.6 \\
        {M-BAB}  & 20.2   & 25.9 & 34.3   & 47.1 \\
        {H-BAB} & 21.8  & 29.2  & 40.1   & 54.1 \\
        \bottomrule
    \end{tabular}
\label{table:UER_final}    
\end{table}

Table~\ref{table:UER_final} shows the average UERs.
The UER was above 10~\% even under the mildest condition (SNR of 15~dB and L-BAB) and increased significantly as the SNR decreased.
This indicates that the GSLM model is vulnerable to noisy inputs when trained with speech without noise. 
We also observed that background speech deteriorated encoding performance, which was further worsened by lower SNR. This observation suggests that the encoder partially encodes background speech.

\smallskip
\noindent \textbf{Phoneme level:} For phoneme-level analysis, we first trained unit-to-phoneme mappings, such as allophone-to-phoneme many-to-one mappings. An allophone is a set of phones mapped to a phoneme in a particular language, e.g., [$\textrm{p}^{h}$, $\textrm{p}$] (phones) $\rightarrow$ /p/ (phoneme). The IBM model 2 aligner with monotonic alignment constraints aligns unit sequences of clean speech and phoneme sequences obtained from texts in the LibriSpeech corpus. We calculated the probability $P\left(\mathrm{phoneme} | \mathrm{unit}\right)$ and mapped each unit to its most probable phoneme, e.g., ["128", "52"] (units) $\rightarrow$ /p/ (phoneme). The phonemizer and aligner used were espeak~\cite{espeak} and \texttt{fast\_align} library~\cite{dyer-etal-2013-simple}, respectively. We calculated the phoneme error rate (PER), just as we calculated the UER in the phone-level evaluation. The reference and hypothesis are phoneme sequences mapped from clean-speech units and noisy-speech units, respectively. Repeated symbols were removed from the sequence to prevent duplication of the same phoneme due to the consecutive estimation of symbols. 
We used 40 phonemes after removing the stress marks from the espeak notation.

\begin{figure}[t]
    \centering 
    \includegraphics[width=0.98\linewidth]{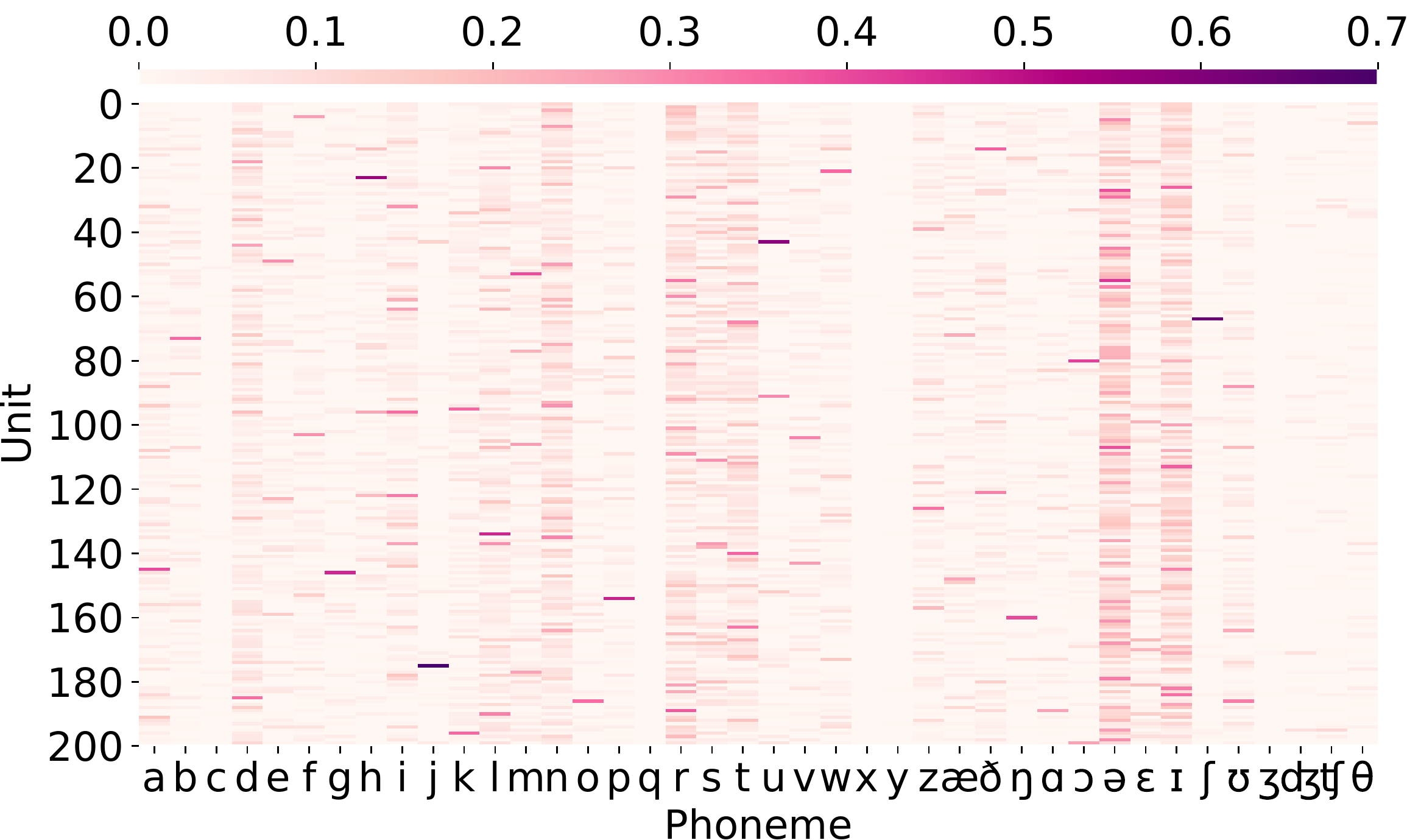}
    \caption{$P\left(\mathrm{phoneme} | \mathrm{unit}\right)$. The darker the color, the more uniquely the unit is mapped by a single phoneme.}
    \label{fig:fig/prob_phoneme_given_unit.pdf}
\vspace{-1mm}
\end{figure}

Before discussing the PER, we show alignments from units to phonemes. Figure~\ref{fig:fig/prob_phoneme_given_unit.pdf} shows the heatmap of the probability $P\left(\mathrm{phoneme} | \mathrm{unit}\right)$. We can find three trends. First, some phonemes, e.g., /g, h, j, u, $\int$/, have high probability to their corresponding phonemes, meaning that these phonemes are more uniquely mapped to the single phonemes. However, /n, r, s, \textipa{@}, I/ have many darker cells and mapped from many units. Finally, /q, x, y, \textipa{\textyogh}/ have only brighter cells. One individual unit is mapped to each of the phonemes, but the probability is low.

\begin{table}[t]
    \centering
    \label{tb:33}
    \caption{PERs [\%] of noisy speech}
    \vspace{-2mm}
    {
        \begin{tabular}{c|cccc}
        \toprule
             & \multicolumn{4}{c}{SNR}\\ 
             & 15~dB &  10~dB & 5~dB & 0~dB \\ 
            \midrule
            {L-BAB} & 14.8 & 18.1  & 21.9 & 27.2 \\
            {M-BAB}  & 20.3   & 25.3 & 32.5   & 43.9 \\
            {H-BAB} & 24.4  & 31.9  & 44.4   & 61.9 \\
            \bottomrule
        \end{tabular}
    }
\vspace{1mm}
\label{table:PER_final}    
\end{table}

Table~\ref{table:PER_final} lists the average PERs with varying SNR. Consistent with the trend shown in UER, PER increased as SNR decreased and reached closer to H-BAB. In M-BAB and L-BAB, lower PERs (Table~\ref{table:PER_final}) were observed compared with UERs (Table~\ref{table:UER_final}). However, the PERs in H-BAB increased regardless of the SNR. Therefore, we can deduce the following two points: First, some unit errors in L-BAB and M-BAB are within allophones and negligible at the phoneme level, and second, unit errors in H-BAB change the phonemes. This is reasonable if the background speech in H-BAB causes changes in the phoneme level.

\begin{figure}[t]
    \centering 
    \includegraphics[width=0.80\linewidth]{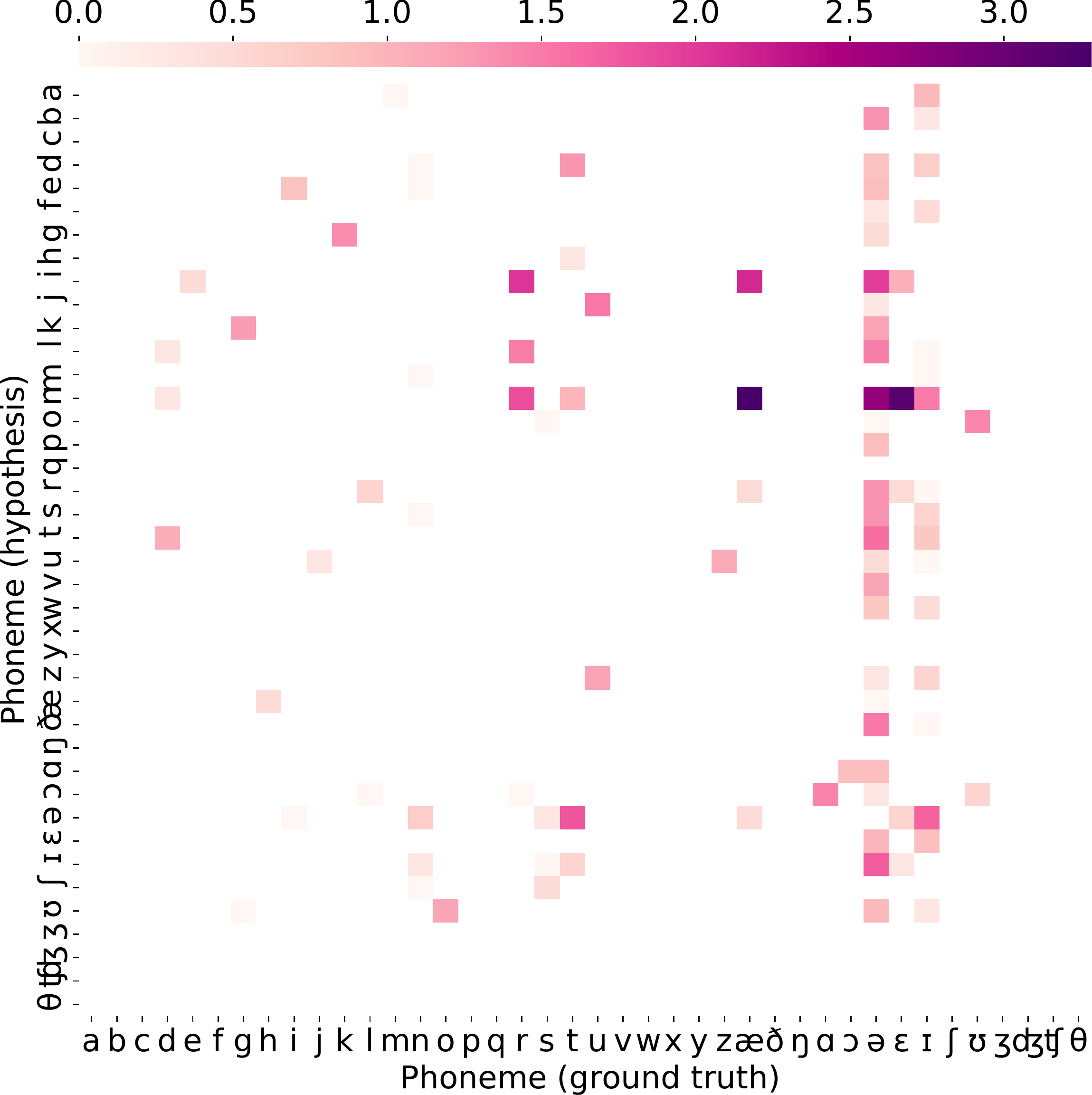}
    \caption{Heatmap of confusion matrix at phoneme level. Values are on log10 scale. Only misrecognized phonemes are counted; correctly recognized ones are not shown in this figure. Noise condition is mildest case: 15~dB SNR of L-BAB noise.}
    \label{fig:fig/phoneme_confusion_matrix.pdf}
\vspace{-1mm}
\end{figure}

To discuss the PER results, we computed the confusion matrix of the phonemes. Specifically, we calculated the matrix by using \texttt{fast\_align} to align the groundtruth and hypothesis phonemes used in the PER calculation. The results of the confusion matrix are shown in Figure~\ref{fig:fig/phoneme_confusion_matrix.pdf}. First, focusing on the groundtruth phonemes (i.e., x-axis), we see that most of the errors are concentrated in specific phonemes, e.g., /e, \ae, r/, etc. Therefore, we can say that units corresponding to these phonemes entangle the phonemes. Next, focusing on the error pairs (i.e., the entire matrix), the matrix is very sparse. Some errors are close in articulation, e.g., /r/ (x-axis) $\rightarrow$ /l/ (y-axis); however, most errors are not, e.g., vowels $\Leftrightarrow$ consonants. Therefore, it is likely that the GSLM encoder captures features that are different from the human articulations.

\smallskip
\noindent \textbf{Word level:} For the word-level analysis, we compared the noisy and resynthesized speech in terms of WER. The references are transcriptions provided by the LibriSpeech corpus, while the hypotheses are automatic speech recognition (ASR) results obtained from the noisy/resynthesized speech by using Whisper-base~\cite{radford2022whisper}.

\begin{table}[t]
    \centering
    \caption{WERs [\%] of raw input (before slash) and resynthesized (after slash) speech per SNR.
    Values of CLEAN are for clean speech}
    \vspace{-2mm}
    {
        \begin{tabular}{c|cccc}
        \toprule
            & \multicolumn{4}{c}{SNR} \\
            & 15~dB &  10~dB & 5~dB & 0~dB \\ 
            \midrule
            {CLEAN} & \multicolumn{4}{c}{4.3/14.5}\\
            \midrule
            {L-BAB} & 5.4/15.3 & 5.5/16.5  & 5.4/20.3 & 7.9/27.1 \\
            {M-BAB}  & 5.4/17.1  & 6.0/20.8 & 7.7/30.9   & 12.8/52.5\\
            {H-BAB} & 5.8/19.8   & 8.5/28.6  & 11.4/49.0   & 25.5/75.8 \\
            \bottomrule
        \end{tabular}
    }
\label{table:WER_final_fix}    
\end{table}

Table~\ref{table:WER_final_fix} lists the average WERs of the noisy and resynthesized speech. Since Whisper had recognition errors on some clean speech, we computed WERs for the clean speech and their resynthesized versions for reference. Comparing the input and resynthesized speech, the difference in WER was above ten points ($4.3 \rightarrow 14.5$) for the reference. Compared with the reference, the WERs for the noisy inputs increased only a few points in L-BAB and at the SNRs of 10 and 15~dB. The results indicate that GSLM resynthesis dominates in word-level degradation rather than mild non-speech noise contamination. 

The WER scores increased significantly under noise conditions with more background speech and lower SNRs, similar to the results in UER and PER. The scores also apparently increased, along with the decrease in SNR and the increase in background speech. These results indicate that noise contamination (particularly, background speech) may change the spoken content in GSLM resynthesis.

\begin{figure}[t]
    \centering 
    \includegraphics[width=0.98\linewidth]{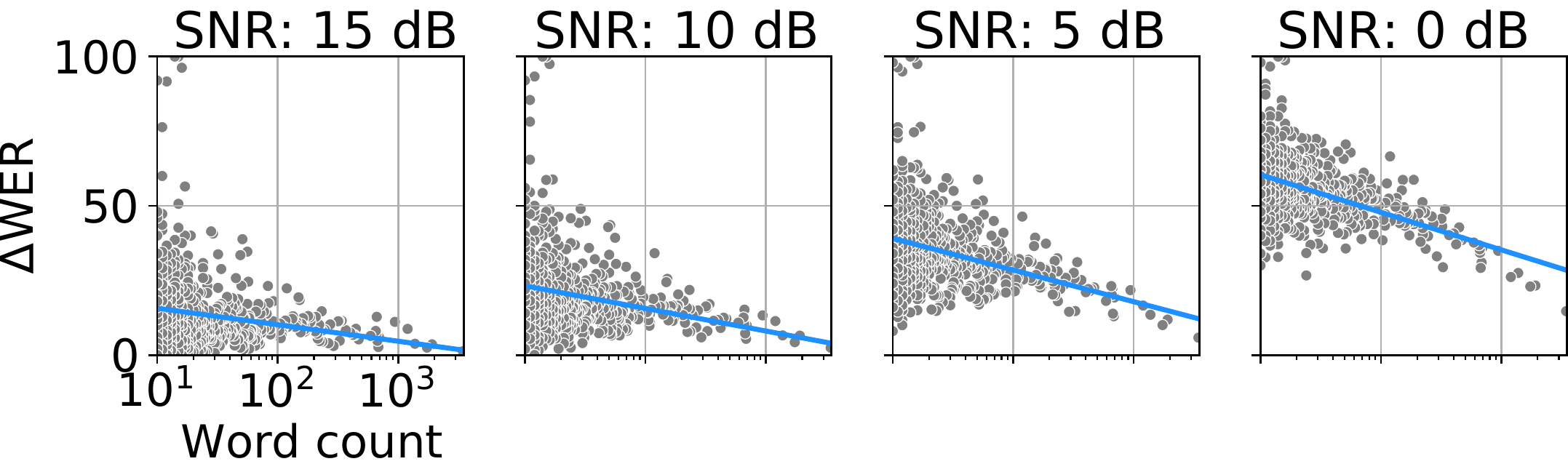}
    \caption{Scatterplots between word frequency and $\Delta$WER per SNR. Blue lines are regression lines.}
    \label{fig:fig/ERWD_final_fix3.pdf}
\vspace{-1mm}
\end{figure}

For a more detailed analysis, we investigated the relationship between word usage frequency in the corpus and WER.
To remove the effects of speech recognition performance, we defined a normalized WER ($\Delta$WER) by subtracting the WERs of the CLEAN noise category from those of L-BAB, M-BAB, and H-BAB for the resynthesized speech.

Figure~\ref{fig:fig/ERWD_final_fix3.pdf} shows scatterplots between word frequencies and $\Delta$WER per SNRs.
The $\Delta$WERs were negatively correlated with word frequency, which became more pronounced as SNR decreased.
This result indicate that word substitutions are more likely to occur for less frequent words; 
for example, words such as ``the'' appeared the most in the corpus and showed a WER of 3.3~\%, but words such as ``ill'' and ``law'', which have the same number of phonemes appeared about $100$ times in the entire corpus, and showed a WER of around 40~\%.
We can also observe that both the slope and bias of the regression line increase as the SNR decreases. This indicates that noise contamination not only degrades the overall performance of word recognition but also worsens the recognition of rare words.

\smallskip
\noindent \textbf{Syntactic level:} To quantify syntactic errors by resynthesis, we used a WER counterpart for word class, i.e., part-of-speech, which we call the word class error rate (WCER).
The references and hypotheses are the word class sequences extracted from the texts of the LibriSpeech corpus and the speech recognition results, respectively.
The word classes were estimated using the natural language toolkit~\cite{bird2009natural}.

\begin{table}[t]
    \centering
    \caption{WCERs [\%] of raw input (before slash) and resynthesized (after slash) speech per SNR.
    Values of CLEAN are for clean speech}
    {
        \begin{tabular}{c|cccc}
        \toprule
            & \multicolumn{4}{c}{SNR} \\ 
            & 15~dB &  10~dB & 5~dB & 0~dB \\ 
            \midrule
            {CLEAN} & \multicolumn{4}{c}{3.1/11.7}\\
            \midrule
            {L-BAB} & 4.0/12.3 & 4.0/13.1  & 4.5/16.1 & 5.7/21.6 \\
            {M-BAB}  & 3.9/13.6  & 4.4/16.5 & 5.8/24.4   & 9.9/41.8\\
            {H-BAB} & 4.3/15.6   & 6.5/23.1  & 8.8/39.6   & 20.6/61.8 \\
            \bottomrule
        \end{tabular}
    }
\vspace{-1mm}
\label{table:POSER}    
\end{table}

Table~\ref{table:POSER} lists the average WCERs of the noisy and resynthesized speech per SNR.
We observe a similar trend in WCER as in UER, PER, and WER.
Regardless of the SNRs and noise category, the WCERs were slightly lower than the WERs but still large.
This result indicates that the word substitution caused by the resynthesis from the noisy speech frequently changes the word class.

\subsection{Analysis at speech level}\label{ssec:quality}
We next evaluated GSLM at speech levels by quantifying the resynthesis quality of GSLM under noisy environments.
For this evaluation, metrics for measuring codec distortion and resynthesized speech quality were used. The former metric is within speech levels, while the latter is a comparison between speech and spoken language levels.

\smallskip
\noindent \textbf{Codec distortion:} Since the discrete symbol sequence is a compressed speech representation, GSLM can be viewed as a neural speech codec. Thus, we investigated the distortion between an input speech and its resynthesized version, which we call \emph{codec distortion}. The codec distortion was measured using WARP-Q, which was developed to predict the quality of neural speech codecs~\cite{Wissam_IET_Signal_Process2022}. It internally compensates for the temporal discrepancy between the ground truth and resynthesized signal, and is robust to the changes in pronunciation timing after resynthesis that often occur in neural speech codecs. It ranges from one to five points and increases as the codec distortion decreases. We used clean speech as the ground truths for computing WARP-Q.

\begin{figure}[t]
    \centering
    \includegraphics[width=0.98\linewidth]{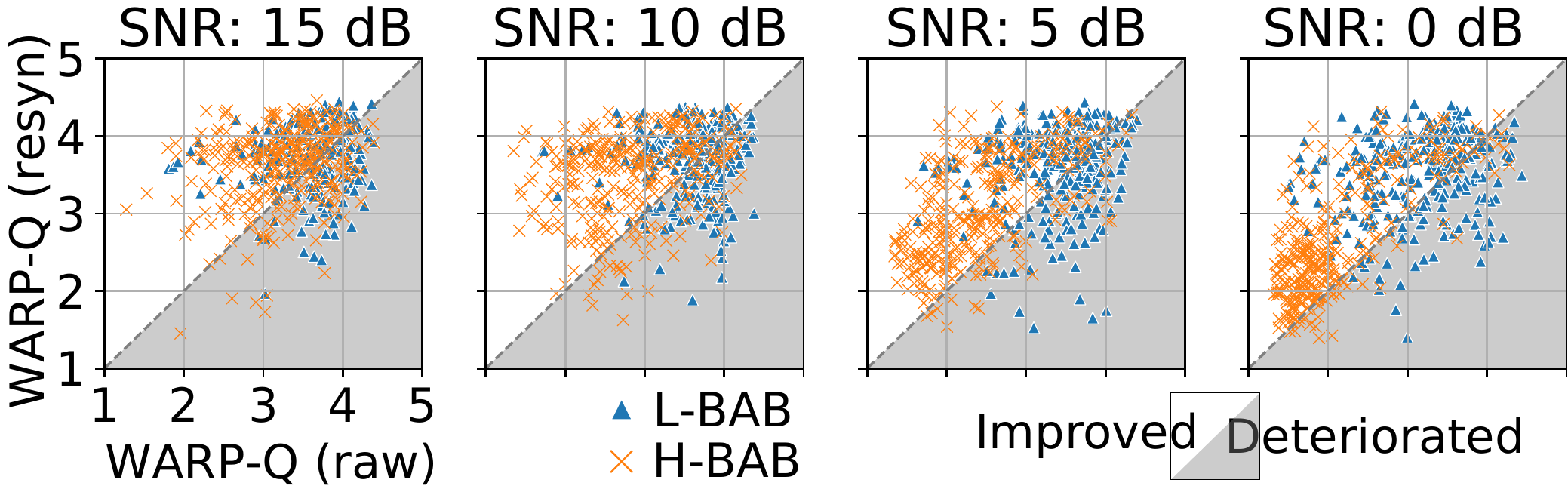}
    \caption{Scatterplots of noisy (raw) and resynthesized (resyn) speech in terms of WARP-Q at various SNRs. Region above diagonal line indicates WARP-Q improvement via GSLM resynthesis.}
    \label{fig:fig/warp-q.pdf}
\end{figure}

Figure~\ref{fig:fig/warp-q.pdf} shows the distributions of the WARP-Q values for noisy and resynthesized speech\footnote{We treated the additive noise as a form of codec distortion and computed the WARP-Q values for the noisy speech.}. 
The WARP-Q improved from the resynthesis as the SNR decreased and the background speech increased which is consistent with the results in Section~\ref{ssec:intelligibility}.
The WARP-Q values of the noisy speech were weakly or not correlated with those of the resynthesized speech under all noise conditions.
These results indicate that GSLM resynthesis on noisy speech decrease codec distortion when speech is included in the noise, but does not mitigate distortion in general.

\smallskip
\noindent \textbf{Naturalness:} To evaluate the naturalness of the resynthesized speech, we used a pre-trained UTMOS model~\cite{DBLP:conf/interspeech/SaekiXNKTS22} strong learner~\cite{stronglearner}.
It predicts a five-scale pseudo-mean opinion score (MOS) value from a synthesized speech\footnote{Although the UTMOS was trained with English/Chinese synthesized speech of the Blizzard Challenge or Voice Conversion Challenge, and works robustly for speech synthesized in other languages and with speech synthesis systems~\cite{abs-2210-14850}.}.
Unlike WARP-Q, UTMOS does not require clean speech for computation and is agnostic to spoken-linguistic changes between clean and resynthesized speech. Therefore, we investigated the relation between linguistic changes ($\Delta$WER) and speech quality (UTMOS).

\begin{figure}[t]
    \centering
    \includegraphics[width=0.98\linewidth]{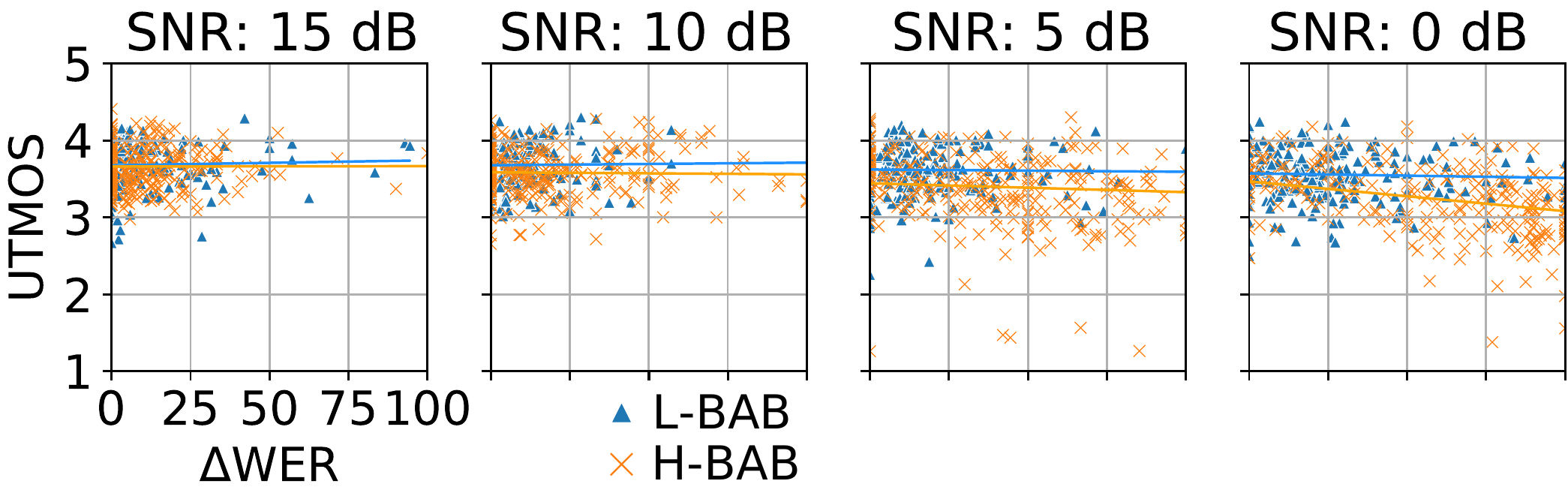}
    \caption{Scatterplots of $\Delta$WERs and MOS values by UTMOS for resynthesized speech. Blue and orange lines are regression lines for {L-BAB} and {H-BAB}, respectively.}
    \label{fig:fig/utmos_scatter.pdf}
\vspace{-1mm}
\end{figure}

Figure~\ref{fig:fig/utmos_scatter.pdf} shows the sentence-by-sentence relationship between $\Delta$WER and values by UTMOS.
The values by UTMOS were distributed around 3.8 points, and their average decreased as SNR decreased.
This result is consistent with the observation in the literature where the discrete speech representation is applied to speech separation/enhancement \cite{2112.09382}.
There were either weak correlations or none between $\Delta$WER and UTMOS values under all noise conditions, showing that GSLM frequently converts noisy speech into natural but content-altered speech. L-BAB does not degrade the UTMOS value but worsens the WER.

\vspace{-2mm}
\section{Conclusion}
We assessed the GSLM performance for noisy speech at the spoken-language and speech levels through speech-resynthesis experiments. The analysis at the spoken-language levels shows that the noise contamination strongly affects resynthesis performance and promotes frequent changes in phones, phonemes, words, and word classes. 
It also suggests that the GSLM encoder captures different speech features from human articulations. The analysis at the speech levels reveals that GSLM frequently resynthesizes natural but content-altered speech and has the risk of falsification if GSLM is used in real (noisy) situations without any special care. 

\smallskip
\footnotesize{
\noindent \textbf{Acknowledgments.} This work is supported by JSPS KAKENHI 22H03639, 23H03418, and Moonshot R\&D Grant Number JPMJPS2011. 
}

\clearpage
\bibliographystyle{IEEEtran}
\bibliography{mybib}

\end{document}